\documentclass[a4paper, 10 pt, conference]{ieeeconf}

\IEEEoverridecommandlockouts                              

\usepackage{xcolor}

\definecolor{MyDarkBlue}{rgb}{0,0.08,1}
\definecolor{MyDarkGreen}{rgb}{0.02,0.6,0.02}
\definecolor{MyDarkRed}{rgb}{0.8,0.02,0.02}
\definecolor{MyDarkOrange}{rgb}{0.40,0.2,0.02}
\definecolor{MyPurple}{rgb}{111,0,255}
\definecolor{MyRed}{rgb}{1.0,0.0,0.0}
\definecolor{MyGold}{rgb}{0.75,0.6,0.12}
\definecolor{MyDarkgray}{rgb}{0.66, 0.66, 0.66}

\usepackage{graphics} 
\usepackage{epsfig} 
\usepackage{mathptmx} 
\usepackage{times} 
\usepackage{amsmath} 
\usepackage{amssymb}  
\usepackage{color}
\usepackage{gensymb}
\usepackage{siunitx}

\newcommand{\myparagraph}[1]{\vspace{0.1in}\noindent\textbf{#1}}

\title{\LARGE \bf GelSight Fin Ray: Incorporating Tactile Sensing into a Soft Compliant Robotic Gripper

}

\author{
    \authorblockN{Sandra Q. Liu and Edward H. Adelson}
        \authorblockA{Massachusetts Institute of Technology\\
    {\tt\small sqliu@mit.edu, adelson@csail.mit.edu}} 

}

\usepackage{multirow}
\usepackage{comment}

\begin{document}

\maketitle
\thispagestyle{empty}
\pagestyle{empty}

\begin{abstract}

To adapt to constantly changing environments and be safe for human interaction, robots should have compliant and soft characteristics as well as the ability to sense the world around them. Even so, the incorporation of tactile sensing into a soft compliant robot, like the Fin Ray finger, is difficult due to its deformable structure. Not only does the frame need to be modified to allow room for a vision sensor, which enables intricate tactile sensing, the robot must also retain its original mechanically compliant properties. However, adding high-resolution tactile sensors to soft fingers is difficult since many sensorized fingers, such as GelSight-based ones, are rigid and function under the assumption that changes in the sensing region are only from tactile contact and not from finger compliance. A sensorized soft robotic finger needs to be able to separate its overall proprioceptive changes from its tactile information. To this end, this paper introduces the novel design of a GelSight Fin Ray, which embodies both the ability to passively adapt to any object it grasps and the ability to perform high-resolution tactile reconstruction, object orientation estimation, and marker tracking for shear and torsional forces. Having these capabilities allow soft and compliant robots to perform more manipulation tasks that require sensing. One such task the finger is able to perform successfully is a kitchen task: wine glass reorientation and placement, which is difficult to do with external vision sensors but is easy with tactile sensing. The development of this sensing technology could also potentially be applied to other soft compliant grippers, increasing their viability in many different fields.

\end{abstract}

\section{INTRODUCTION}


Compliant and soft robotic fingers are capable of providing both robust adaptability for grasping a multitude of different objects as well as safety for day-to-day interaction with humans. Due to their material properties and passivity, they are able to easily grasp fragile and soft objects, can accommodate various-sized objects with unique shapes, and do not require energy-extensive actuation. Overall, they provide more universality in their grasping capabilities compared to their rigid, power-expensive robotic counterparts and can be more robust and durable. 

Recent progress in soft, compliant robotics has enabled an increase in the development of safe and reliable human-robot interactions, with specific focus towards agricultural purposes, robotic surgery, prosthetics, and aiding the elderly to age with dignity \cite{biomedical, agriculture, hri}. Many of these fields require robots to have a degree of compliance, similar to the compliance human hands have, whether it is the ability to use less force when interacting with soft, ripe fruit or helping a human without accidentally hurting them in the case of a system failure. In particular, as the average age of the human population increases, it is important to develop a safe, but extremely capable, home robotic assistant that can assist with chores in the kitchen. One such task is the ability to reorient dishware (i.e. wineglasses) and safely place them on a table so that they do not break, which could cause harm to others. 

\begin{figure}[t]
	\centering
	\includegraphics[width=1.0 \linewidth]{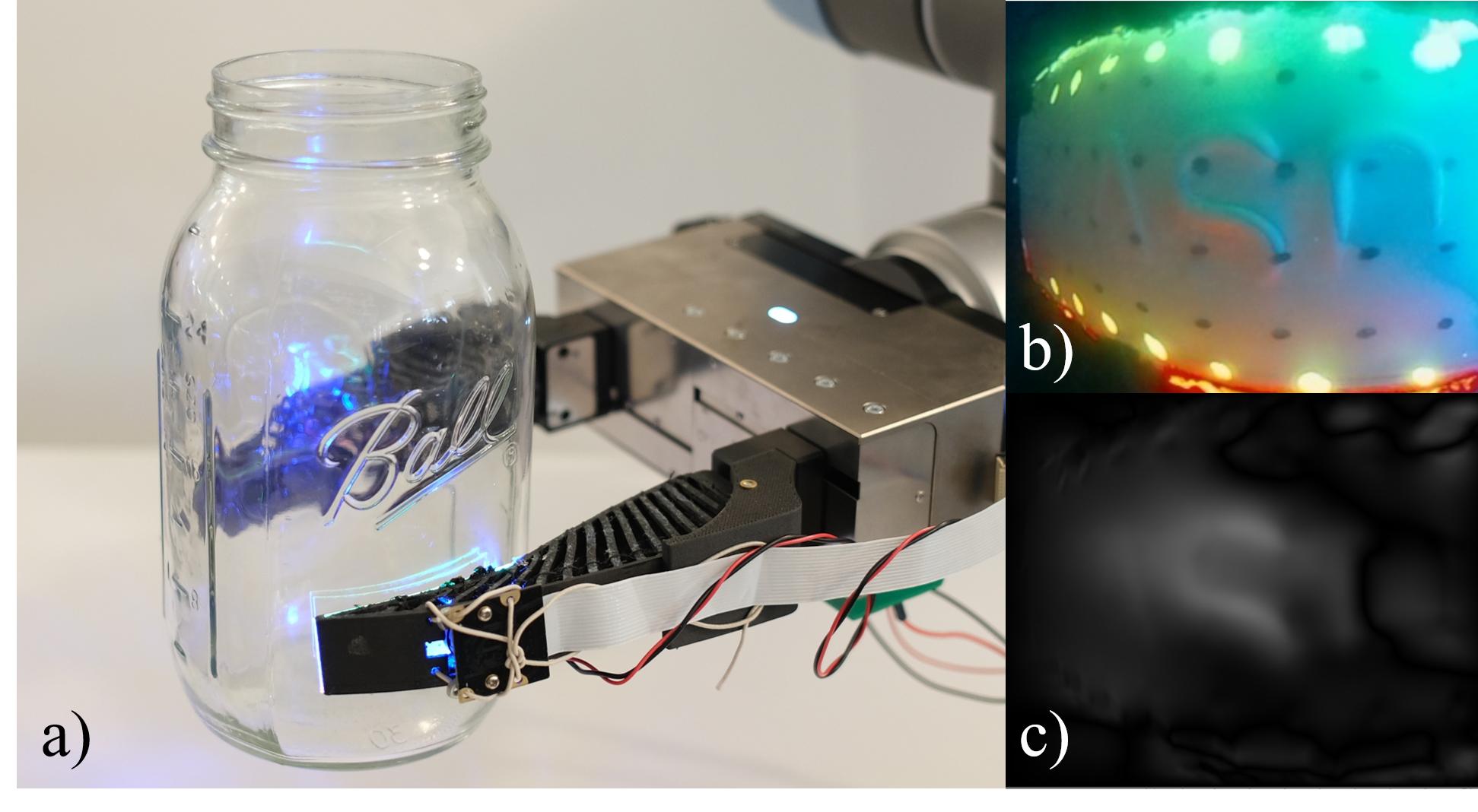}
    \vspace{-20pt}
	\caption{ In a), the GelSight Fin Ray gripper is holding a ball glass mason jar with its tactile sensing region along the indented letters ``MASON" while b) denotes the 90$^\circ$ counterclockwise-rotated raw tactile image data where the top of the image generally denotes the tip of the finger and c) is its corresponding uncalibrated tactile reconstruction, where you can see most of the ``S" letter and parts of the ``O" and ``A" letters as well.}
	\label{fig:bigpic}
	\vspace{-10pt}
\end{figure}

For a robotic gripper to have such capabilities, it is not sufficient to only be compliant. Robotic grippers must also be able to utilize sensing, such as tactile sensing, which can aid in manipulation \cite{wang2018toward}. Tactile sensing can provide information about surface roughness, object orientation, contact force, and slip measurements \cite{tactilesensing}. Specifically, this type of sensing can more effectively determine how a soft compliant gripper is grasping an object and reorient it so that it can be safely placed on a table without a catastrophic failure. Having tactile sensing is essential for building at-home robots that can safely aid the elderly and improve their overall quality of life as they age. It is important to combine the adaptive advantages of a soft compliant gripper with the general-usage manipulation abilities that is provided by tactile sensing.

This paper presents the following contributions:
\begin{itemize}
    \item A novel design of a fully-functioning flexible, elongated tactile sensing surface with an embedded vision sensor for tactile reconstruction, orientation estimation, and slip detection;
    \item A novel design of a sensorized GelSight Fin Ray inspired gripper with tactile sensing (Fig. \ref{fig:bigpic});
    \item An algorithm that allows the gripper to be handed a thin object, determine its orientation, and gently set it down upright using only tactile sensing. 
\end{itemize}

\section{RELATED WORK}
\subsection{Compliant/Fin Ray Inspired Grippers} \label{review1}

Recent work in passive adaptive grippers has led to the development of various iterations of a 3D printable Fin Ray Effect gripper, which has the ability to gently comply to an object it comes in contact with via the fin-like structure of the fingers \cite{festo}. The idea was originally inspired by the study of fish fin bone structure and its passive, deformable geometry \cite{biofinray}. This geometry allows Fin Ray fingers to conform to a surface, increasing the grasping contact surface area to allow a more secure grasp \cite{sorogrip, ninjaflex}. These types of fingers, unlike other soft robotic fingers, are relatively simple to manufacture and easy to modify to incorporate various materials or control schemes \cite{finrayog}. Furthermore, their design does not require actuation to securely grasp objects, unlike many other rigid and soft robotic grippers, which allows Fin Ray fingers to potentially be used in cleaner or limited energy environments \cite{manta}. These qualities are important for performing manipulation in an unstructured home space, and as a result, a lot of research has been done to optimize the Fin Ray finger design. 

To streamline the Fin Ray struts and its design to more optimally grasp a large variety of objects, both Shan et al. and Deng et al. developed mathematical models and simulations to assess and optimize its grasping capabilities \cite{finrayshan, windingdeng}. Crooks et al. worked on adapting the original Fin Ray design to ensure a preferred bending direction and increasing the grip force of the fingers \cite{finrayog} while Elgeneidy et al. worked on improving the design of a NinjaFlex Fin Ray finger through experimental design testing methods \cite{experimentalninflex}. Another solution to improve grasp quality of a Fin Ray gripper involved adding electroadhesion pads to the tactile surface of the fingers \cite{electro}.

Despite the progress made towards optimizing the Fin Ray finger design, not much work has been done to sensorize it. Although Yang et al. proposed a Fin Ray finger that incorporates an embedded force sensor for detecting the finger state and to measure contact force \cite{finrayforce}, it does not provide the higher-detailed tactile sensing necessary for more complex manipulation tasks. 

\subsection{Soft Robot Tactile Sensing} \label{review2}

In general, most of the work done with soft robotic tactile sensing has been through strain sensors \cite{tendon, soter2018bodily}, which are only able to provide information on force contact and cannot give additional tactile information. To mitigate these shortcomings, vision-based tactile sensing can be used instead. 

Embedded tactile sensing via camera sensors can provide high-resolution information of the contact area that will not be obscured by outside elements. However, the majority of vision-based tactile sensors are currently still built only for rigid fingers or fingertips. Even when embedded vision-based sensing is used for soft robotic finger applications, as they are in the work by Hofer et al. and Oliveira et al, the camera is only used to sense proprioception or force deformations \cite{hofer, oliveira}. A vision-based proprioceptive and tactile sensing approach is also used for an exoskeleton-covered soft robotic finger \cite{exoskeletonsoft}. However, the high-resolution tactile sensing provided by the camera is not used to its full potential and cannot provide any tactile reconstruction or marker tracking heuristics that the GelSight sensor can \cite{dong2017improved}.

The GelSight sensor is a vision-based tactile sensor which uses a patterned soft gel as a sensing medium, tri-colored LEDs directed from three different directions, and an embedded camera that tracks the high-resolution tactile information. To the authors' knowledge, currently no other soft robotic gripper has incorporated the full functionality of a GelSight sensor, allowing tactile reconstruction, object orientation detection, and marker tracking. The design introduced in this paper also allows a GelSight sensor to be elongated and flexible, which captures a larger tactile surface area and enables compliant grippers to utilize GelSight sensor technology.

\section{METHODS}

\subsection{Sensor Design} 

\begin{figure}[ht]
	\centering
	\includegraphics[width=1.\linewidth]{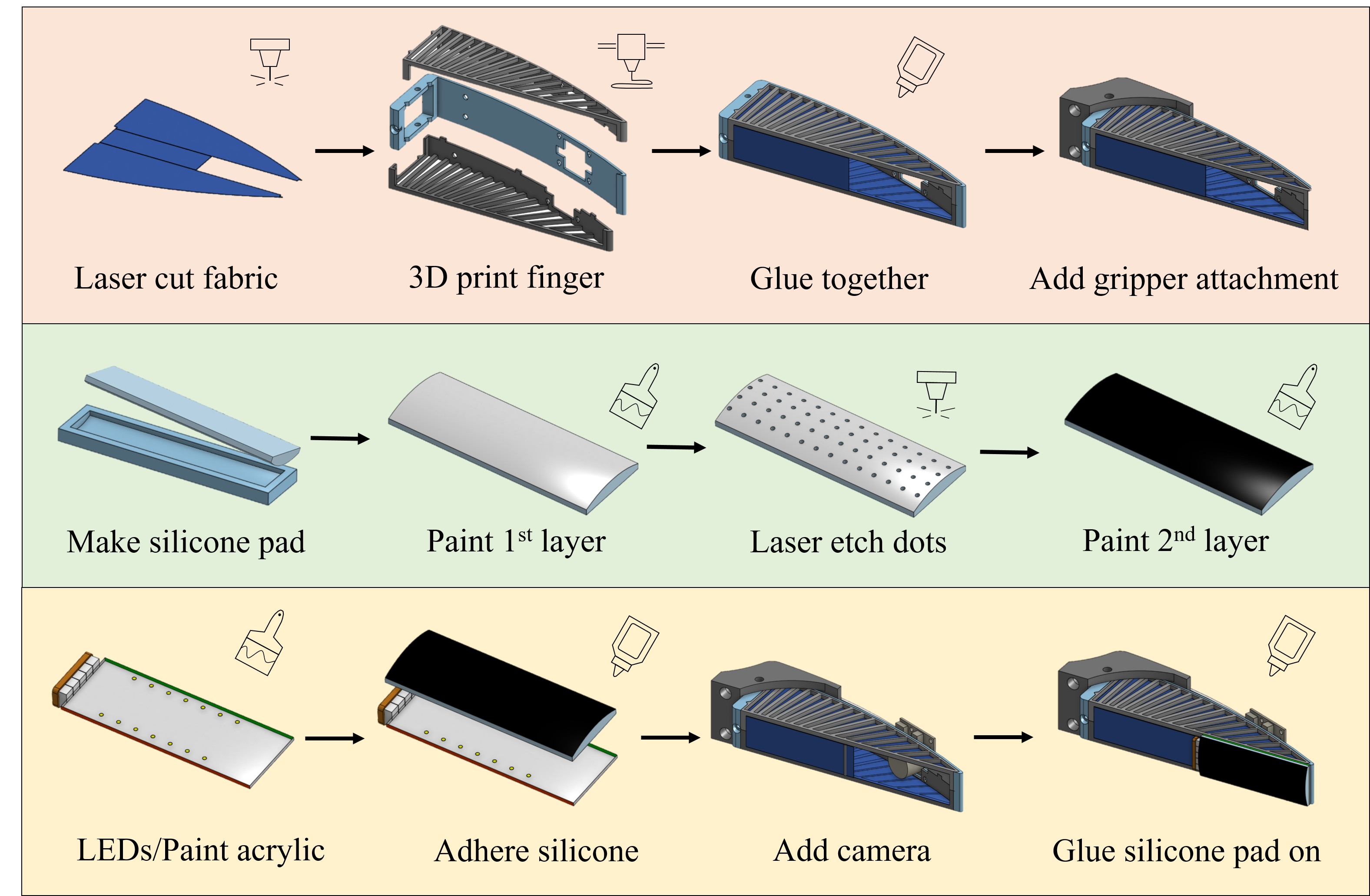}
    \vspace{-15pt}
	\caption{Manufacturing process flow of the GelSight Fin Ray finger, which includes finger design, tactile sensing pad manufacturing, and illumination.}
	\label{fig:manu}
	\vspace{-10pt}
\end{figure}

Overall, the sensor design involves three major components: finger design, tactile sensing pad manufacturing, and illumination. The resulting finger was attached to a parallel gripper configuration. Fig. \ref{fig:manu} shows the entire manufacturing process of the finger.

\myparagraph{Finger Design}

The struts of the finger were designed based on the design proposed by Elgeneidy et al. \cite{lala}. To create a finger that could house and allow usage of a vision sensor while retaining the structural features of a Fin Ray, the middle part of the finger was hollowed out. This removal of the middle support material caused the finger to become too flexible, so a harder material (1.75 mm TPU, Overture, printed on Prusa MK3S) was used for the finger struts while an even more rigid material (Onyx, Markforged) was attached to the back of the finger as support. To circumvent the unwieldy removal of TPU support material from the main finger print, the struts portion of the finger was separated into two parts along its length. Doing so allowed the parts to be printed with the outer section against the build plate, enabling them to be printed with no support material and eliminating extensive post-manufacturing processes while also shortening the printing processes. 

Since the sensor utilizes a camera, it was important that any ambient lighting be obscured so that it did not interfere with the tactile sensor. However, any material added to the finger also had to be stretchy so that it would not interfere with the compliant characteristics of the finger. To solve this issue, a laser cut piece of black, stretchy cloth (spandex/nylon) was adhered to the inside of the struts and across the bottom half of the open part of the finger before the three 3D printed parts were pieced together. The bottom half was covered since most of the manipulation and grasp tasks are performed near the tip of the finger, where the finger is more compliant. The dark cloth obstructed outside environmental lighting while its elasticity allowed the finger to function as it would without the addition of the cloth. Afterwards, the two halves of the finger were glued onto the rigid base with cyanoacrylate glue.

\myparagraph{Tactile Sensing Pad Manufacturing}

The tactile sensing silicone pad was cast as a rectangular-profile piece with dimensions of 2.5 mm by 18 mm, with a 25 mm curved profile along one of the 18 mm surfaces to serve as the tactile sensing surface. The mold was 3D printed using the Markforged Onyx material. To ensure that the tactile curved surface was smooth, a piece of plastic mylar sheet with 6 mil (0.15 mm) thickness was superglued onto the inner curved mold surface. A platinum catalyst translucent silicone (XP-565, Silicones, Inc.) was combined with a plasticizer (LC1550 Phenyl Trimethicone, Lotioncrafter) in a ratio of 11 to 3 (1 part XP-565 activator to 10 parts base to 3 parts plasticizer) to be used for the silicone pad. The mixture was used to ensure a clear, transparent sensing region for the camera sensor and a plasticizer was added to increase the softness and robustness of the silicone pad. After degassing the silicone mixture and pouring it into a pre-prepared mold, the mold was left to cure for 24 hours at room temperature. 

Once the silicone pad was cured, a thin layer of silicone paint was brushed on the curved side of the silicone with a foam brush. This paint layer formed a semi-specular tactile sensing surface. To create this painted surface, which would allow the sensor to more easily pick up small details and dimmer illumination in the tactile sensing surface, a mixture of 1 part silicone ink catalyst to 10 parts gray silicone ink base (Raw Materials Inc.) to 2.5 parts 4 $\mu$m Aluminum cornflakes (Schlenk) to 30 parts NOVOCS Gloss (Smooth-on Inc) was used. After allowing the painted surface to fully cure, 1 mm circular dots spaced 4 mm apart were laser engraved (Epilog Laser Cutter System) on the painted surface, effectively removing the painted layer. A layer of black silicone paint (1 part catalyst to 10 parts black paint to 30 parts NOVOCS Gloss) was then sprayed on using an airbrush, creating patterned black dots on the silver-gray tactile surface. The dots allow for the use of marker tracking to determine motion or slippage of a grasped object. 

One of the manufactured silicone pads was then glued to a piece of 0.5 mm thick acrylic to provide the silicone with a rigid backing and prevent the silicone from folding in on itself when it comes in contact with an object. A mixture of 1 part Silpoxy (Smooth-on Inc) to 10 parts NOVOCS gloss was used so that the silicone could bond to the acrylic. 

The thinness of the acrylic is to ensure that the finger has overall compliance and can still bend when it comes into contact with an object. Acrylic was also chosen as the rigid base of the silicone pad for its ease of manufacturing and its reasonably close index match with silicone, which helps to mitigate light interference of the sensing surface caused by refraction. This acrylic piece was then attached to the 3D printed finger.

\myparagraph{Illumination}

To illuminate the finger, the sides of the acrylic were respectively covered with a layer of red and green fluorescent acrylic paint (Liquitex BASICS ACRYLIC) and a line of blue LEDs (450 nm) was adhered to one end of the acrylic piece to illuminate the silicone pad and cause the paints to fluoresce. A short strip of TPU filament was superglued to its back and along the empty middle section of the finger to hold the blue LEDs in place. Together with the fluorescent paints, the blue LEDs helped to provide a tri-color sensing region, similar to how a GelSight wedge and digger fingers are illuminated \cite{wedge, digger}. The presence of all three colors help the sensor to perform depth reconstruction of objects the gripper comes into contact with.

The camera is set opposite the tactile surface near the tip at the back of the finger so that it can view the most compliant parts of the gripper as well as where the majority of the tactile interactions will occur. To mitigate the saturation of blue LEDs viewed by the camera, a yellow filter is placed atop the camera sensor. The filter helps make the less vivid red and green colors visible by the camera. 

Because illumination of the finger as viewed by the camera changes as the finger flexes, separate yellow fluorescent dots (Liquitex BASICS ACRYLIC) were painted on both sides of the acrylic surface to provide a reference for the shape of the finger. The color was deliberately chosen so that it could easily be visible in the dim illumination and so as not to interfere with the tactile sensing. These dots are separate from the black silicone dots used for marker tracking, which depict the shear information of the silicone surface. 

\myparagraph{Gripper Assembly}

The fully assembled finger was then attached to the parallel WSG 50-110 (Weiss) gripper using an Onyx 3D printed part designed to fit as a replacement for the traditional Weiss gripper fingers.

\subsection{Software}

Similar to a GelSight sensor, the Fin Ray finger is able to perform tactile sensing, measure the orientation of an object it comes into contact with, and track markers for slip and twist detection \cite{dong2017improved}.

All images were captured using a Raspberry Pi 160$^\circ$ field of view (FOV) camera, before being cropped slightly so that only the tactile sensing portion of the image was shown. A wide FOV camera was used so that it can sense along the entire length of the sensing region.

\myparagraph{Tactile Sensing/Reference Image}

The main challenge with incorporating tactile sensing into the Fin Ray finger is that the finger deforms when any object comes in contact with it. As a result, the camera sees both the proprioceptive change when the finger bends, and the minute changes in the silicone pad corresponding to the tactile surface interactions. Tracking only the fine-detailed tactile information requires knowledge of the proprioceptive state of the finger or the bending positions of the acrylic. As such, yellow fluorescent dots were painted on the sides of the acrylic so that the motion of the finger could be tracked and used as a reference. This reference was then used to isolate the small-scale tactile information from the GelSight sensor.

\begin{figure}[ht]
	\centering
	\includegraphics[width=1\linewidth]{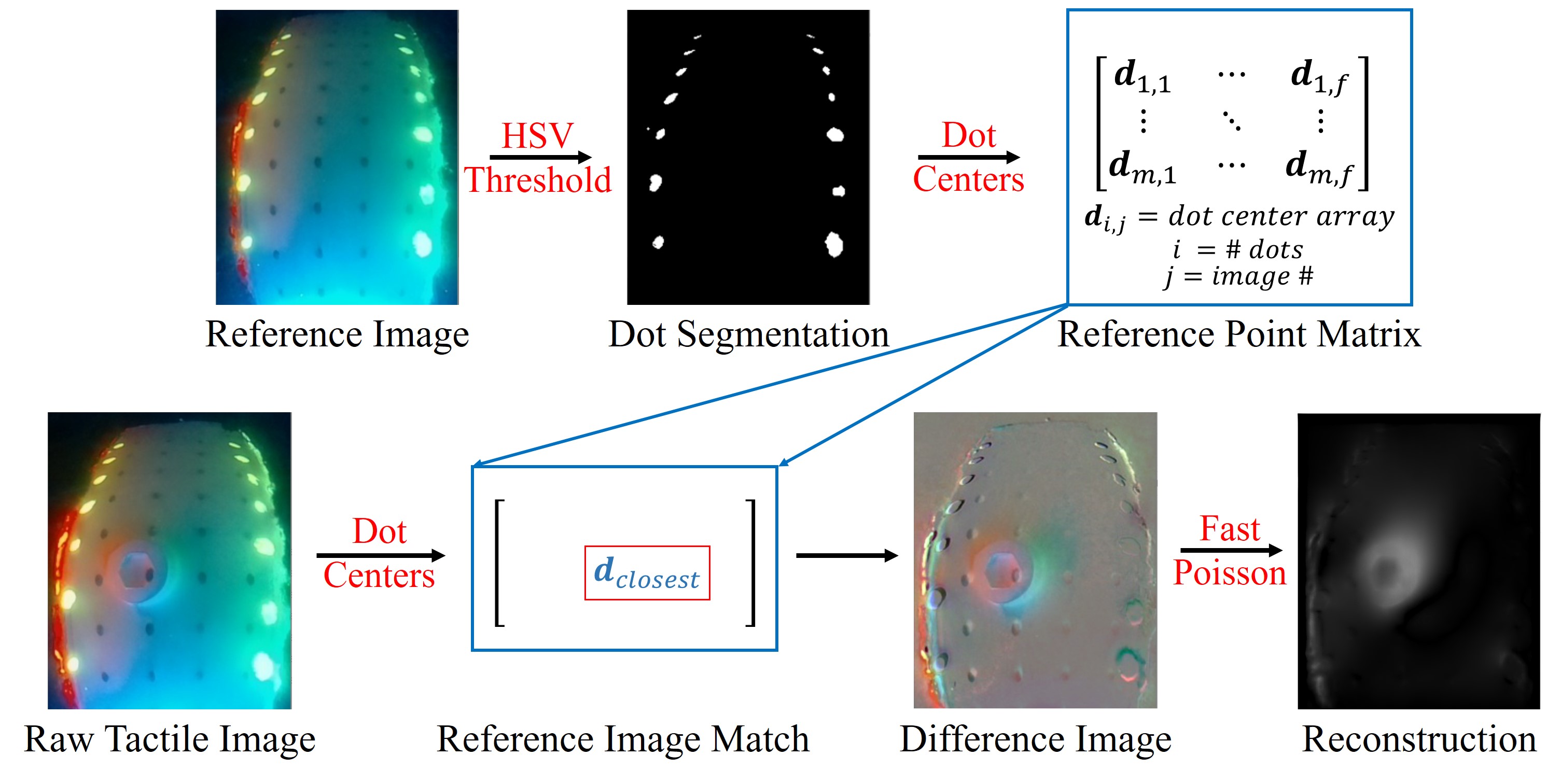}
    \vspace{-15pt}
	\caption{Algorithm flow for obtaining the reference point matrix and live reconstruction of the raw tactile image. The reference point matrix is precalculated using a video taken of the finger without any tactile interference. The matrix is sparse and contains only one nonempty element per column. Subsequent raw tactile images are then compared to the reference point matrix via the dot centers, a difference image is taken, and an uncalibrated reconstruction image is calculated. Pictured in the raw tactile image is the head of a screw that was pressed into the sensing region at a slight angle.}
	\label{fig:algflow}
	\vspace{0pt}
\end{figure}

A reference video was created with the finger where a human hand flexed the sides of the finger which are not visible by the camera sensor. As such, the camera was able to capture the different illumination schemes without any tactile interference. Using the video, the dots were thresholded out via HSV color segmentation, opening and closing operations were performed on the thresholded image, and the individual dot centers were extracted from each frame of the reference videos using a Python OpenCV contour finding algorithm and converted into a Numpy array for better latency \cite{opencv, numpy}. Subsequent tactile images were then compared to reference images using the dot segmentation method to find the image that most closely matched the current image. The general algorithm flow is shown in Fig. \ref{fig:algflow}.

\myparagraph{Object Orientation}

After matching the tactile image with a reference image, difference images were taken and then Poisson reconstruction was used to calculate uncalibrated depth/reconstruction images \cite{poisson}. The contour finding algorithm was then used on thresholded versions of the depth images and principal component analysis was used to find the orientation of the object that was being grasped.

\myparagraph{Marker Tracking}

Marker tracking was performed using the difference image between the luminosity channel of the LAB color space and its corresponding median-blurred image. This method allowed the markers to be segmented out of each image. Arrows were then drawn from the reference image marker to its closest marker in the actual image. If the distance between them was significant, the arrow length was scaled by a factor of three to emphasize the shear and twist motion of the object in contact. 

\section{EXPERIMENT}

\subsection{Experimental Setup}

Software was run via Python and ROS \cite{ROS} and incorporated object orientation estimation and marker tracking of the tactile surface. The camera was run on a Raspberry Pi board and images were streamed to the computer using mjpg\_streamer. The fingers were attached to the Weiss gripper, which was mounted on a UR5 arm. Although both of the fingers in the parallel gripper were Fin Rays, only one of them was sensorized. 

\begin{figure}[ht]
	\centering
	\includegraphics[width=0.8\linewidth]{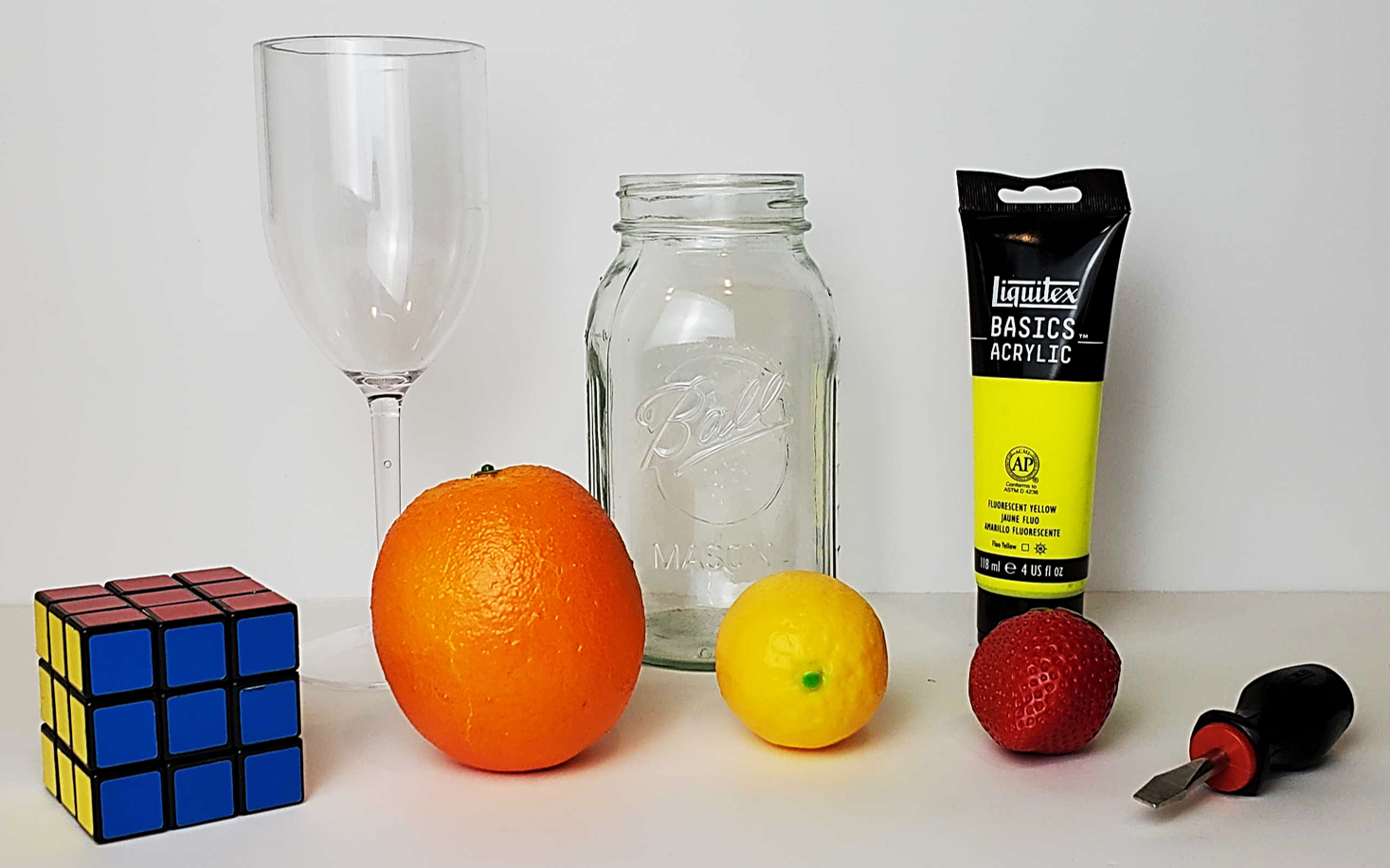}
    \vspace{-5pt}
	\caption{Objects used for the experimental setup. They include a mini screwdriver, plastic fruits, a Rubik's cube, a wine glass, a mason jar, and a squishy acrylic paint tube.}
	\label{fig:objects}
	 \vspace{-5pt}
\end{figure}

To test the compliance and gripping ability of the fingers, many test objects were handed to the gripper to be grasped. These objects included a mini screwdriver, a plastic strawberry, a plastic lemon, a plastic orange, a Rubik's cube, a wine glass, a ball glass mason jar, and a deformable, squishy acrylic paint tube (Fig \ref{fig:objects}).

To test the orientation estimation and marker tracking, the gripper was programmed to be handed a wine glass stem, measure its orientation based on tactile results, and rotate the gripper so that the wine glass would be oriented upright. Afterwards, the gripper would set the wine glass down until the total magnitude, or shear force, detected by the marker tracking algorithm exceeded a threshold, indicating that the glass had touched the table. A wine glass was chosen because it is difficult to segment transparent objects using vision, and thus, tactile sensing greatly simplifies the reorientation and placement problem.

\subsection{Results}

\myparagraph{Grasping} 

The Fin Ray fingers were able to comply to and grasp all of the test objects, which included a variety of different sizes and shapes. For curved objects, the Fin Ray fingers were able to wrap around the curvature of the grasped objects while for the more linear objects, such as the screwdriver and the Rubik's cube, the fingers acted more like a rigid, parallel gripper. For the squishy paint tube, the Fin Ray fingers were able to comply around the object without breaking or collapsing the paint tube. Although the grips for heavy and asymmetrically weighted objects were subject to slippage, the grasp became sturdier once the force the gripper could apply was increased.

This difficulty in grasping particular objects may be partially due to the slight inadvertent twisting motion that the fingers sometimes exhibited due to the hollowing out of the structure inside, which most likely decreased the gripping strength of the fingers. Despite the addition of the acrylic mount, which did limit the twisting motion, there was still a minute amount of twist in the finger itself. Furthermore, the acrylic piece could have also created challenges in grasping since it may have limited the flexibility of the finger and required more force to ensure greater compliance of the finger to the object it was grasping. There was a tradeoff between the twisting motion of the structure and its compliant flexibility. However, this tradeoff did not result in hindering any of the compliance or grasping abilities of the Fin Ray finger for the objects in the testing set.

In summary, the GelSight Fin Ray gripper was able to grasp a multitude of objects, showcasing that the fingers still retain their compliance properties, which makes them useful for a large variety of grasping tasks that potentially need robust, universal grasping abilities combined with tactile sensing. 

\myparagraph{Tactile Sensing}

\begin{figure}[ht]
	\centering
	\includegraphics[width=0.8\linewidth]{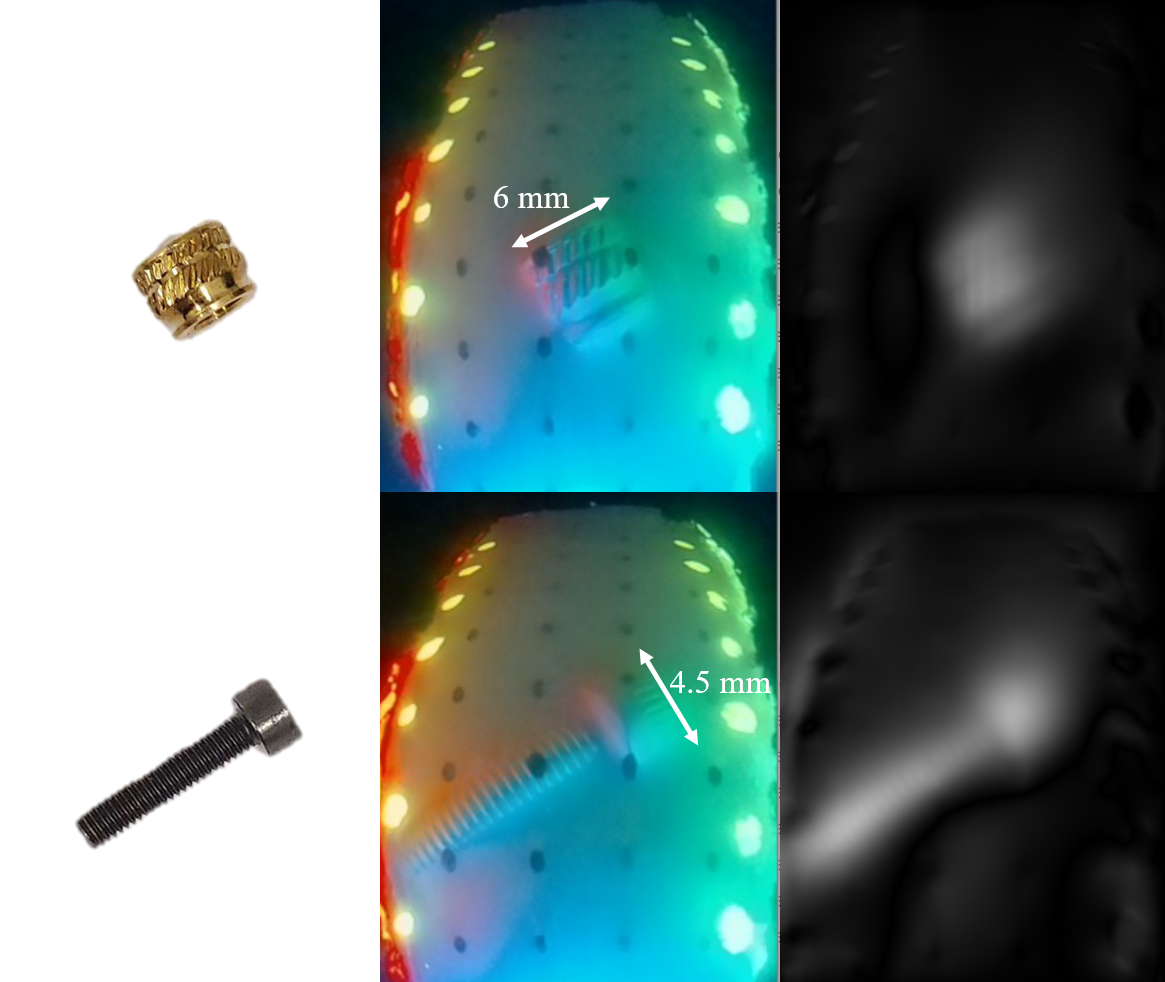}
    \vspace{-8pt}
	\caption{Tactile sensing image reconstruction. The left most images show the objects (M4 heated insert, M2.5 screw) that were placed on the tactile sensing images, the middle images represent the raw tactile data, and the right most images display the reconstruction based on the raw data.}
	\label{fig:tactile}
	 \vspace{-15pt}
\end{figure}

The tactile surface of the sensor is able to detect minute details, such as some of the threads of a M2.5 screw, and the indents on the outside of a M4 heated insert. All of the reference images were able to help provide good reconstruction images, despite the slightly nonuniform fluorescing quality of the red and green paints. The raw tactile images and reconstruction images are shown in Fig. \ref{fig:tactile}. 

\begin{figure}[ht]
	\centering
	\includegraphics[width=0.65\linewidth]{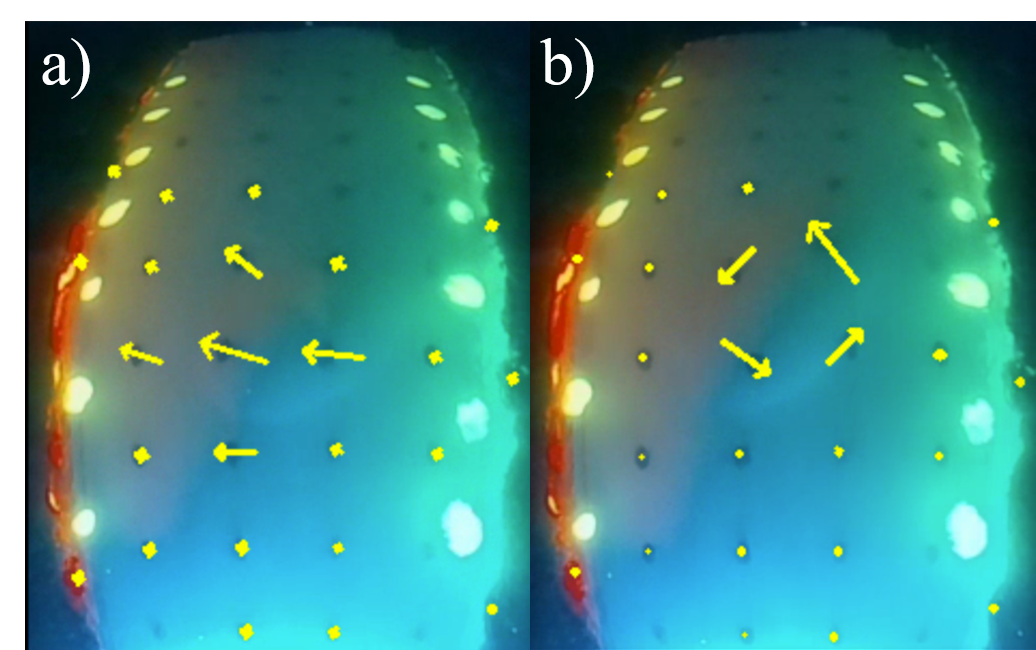}
    \vspace{-3pt}
	\caption{Marker tracking with the GelSight Fin Ray. The yellow arrows denote the tracked position from the reference markers in the found reference image to the markers in the current image. In both images, the fingers are gripping a smooth acrylic cylinder. The left image (a) denotes shear along the cylinder from an external force while the right image (b) displays torsional external force being applied on the cylinder.}
	\label{fig:stocker}
	 \vspace{-5pt}
\end{figure}

Generally, the dot center matching algorithm was quite robust, but it seemed to struggle more when the GelSight Fin Ray fingers wrapped around heavier objects such as the large mason jar. The extra weight caused the fingers to twist a bit and would shift the raw tactile image down by a nontrivial amount, causing some noise in the uncalibrated reconstruction image since it was difficult to include twisting motion in the reference images without interfering with the tactile sensing region. Regardless, the marker tracking was able to sense the shear force motion due to twisting by gravity and was able to sense twist and shearing motions along the center of the tactile surface (Fig. \ref{fig:stocker}). 

Due to the proximity of the black dots to the fluorescent dots, there were cases where some of the black dots could not be perfectly tracked. This phenomena also occurred for the dots in some of the regions at the top of the raw tactile image, which are furthest from the illumination of the blue LEDs. Having black dots also made the algorithm track a few black spots at the outer edges of the tactile sensing, which served to add some noise into the system as well. Nevertheless, all of this noise did not affect the marker tracking overall, and can be potentially optimized for future design iterations of the GelSight Fin Ray.

\myparagraph{Wine Glass Reorientation}

\begin{figure}[ht]
	\centering
	\includegraphics[width=1.0\linewidth]{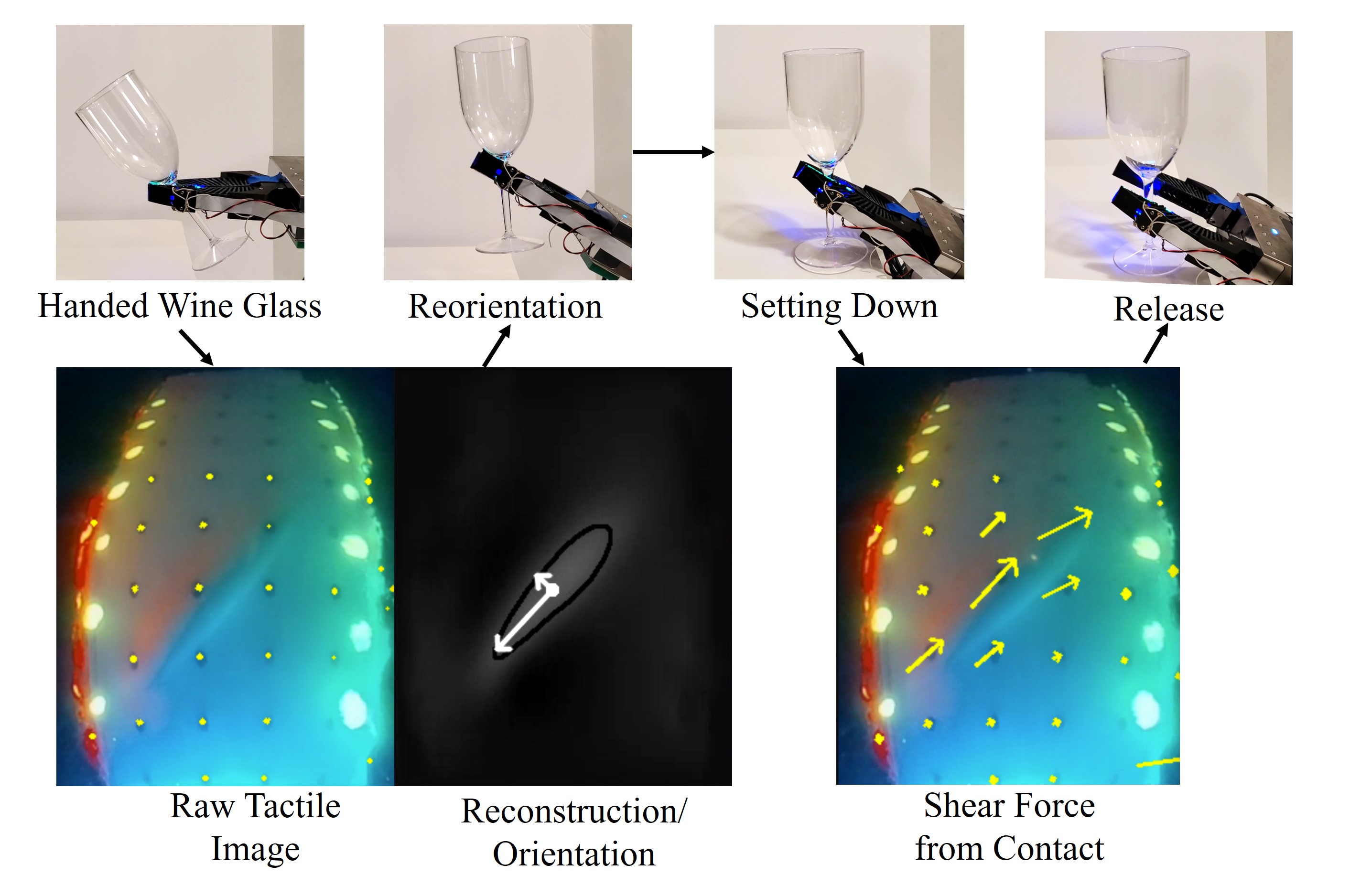}
    \vspace{-20pt}
	\caption{Successful implementation of wine glass reorientation. A wine glass is handed to the gripper at an angle. Using the raw tactile image, the orientation of the wine glass stem is found through the reconstruction algorithm. Using this angle, the UR5 arm reorients the gripper so that it is holding the wine glass right side up. The arm then sets the wine glass down until the shear force from the contact goes above a certain threshold and causes the gripper to release the wine glass on the table without it tipping over. Note that the shear force detected by the marker tracking algorithm is along the major direction of the stem, indicating that the wine glass is upright.}
	\label{fig:vinovino}
	 \vspace{0pt}
\end{figure}

Without tactile sensing, the task of reorienting and setting down a wine glass without tipping it over or crashing it into the table becomes more complex due to its transparency, which makes it harder to see. With the GelSight Fin Ray, however, wine glass reorientation was generally successful (Fig. \ref{fig:vinovino}). Out of the 10 trials that were performed, the algorithm was able to succeed in 7 of them; the GelSight Fin Ray finger was able to allow the gripper to successfully reorient and set the wine glass down without it tipping over. In the three cases where the algorithm failed, the grasp was not secure enough and the wine glass slightly slipped out of the Fin Ray grasp, causing the reconstruction of the image to fail, which in turn caused failure of the reorientation portion of the wine glass. However, in those cases, the gripper was still able to detect when the wine glass came in contact with the table and stop itself from crashing the wine glass into the table further and potentially breaking it. 

To conclude, the marker tracking portion of the software, despite some noise in the system, was more robust than the reconstruction algorithm. This robustness most likely occurs because the marker tracking algorithm does not require firm contact of the silicone pad sensing surface with its object. Bare minimum contact is needed as long as the top part of the silicone pad can be visibly displaced by external forces. Since the silicone pad is soft, the top portion of the pad was easily deformable, causing the marker tracking portion of the tactile sensing to be inherently more robust. The future addition of a neural network could also generally increase the accuracy and results of tactile reconstruction and marker tracking. 

\section{CONCLUSION AND DISCUSSION}
State estimation via tactile sensing is an important issue in soft robotic hand manipulation. By better understanding and being able to perceive their own state and outside tactile stimuli, soft compliant robots may increase their ability to interact with the environment around them and help increase the quality of life of the people they are trying to aid.

The incorporation of tactile sensing also has the potential to allow soft compliant robots to mimic or outperform human capabilities to reconstruct the world around them through tactile manipulation. Such abilities may ultimately allow soft robots to improve their performance in medical fields (i.e. surgical robots or prosthetics), during human-robot interactions, or on rescue missions in unknown situations. In particular, Fin Ray fingers allow manipulation to be performed in limited energy environments due to their simplicity in actuation, enabling them to be of great usage in many different environments around the world. However, many existing sensors that provide very intricate tactile information are rigid, compact, and not suitable for utilization in a soft robot.

This paper introduced a novel flexible, elongated GelSight sensor that can be used in conjunction with a Fin Ray inspired finger, allowing it to perform reconstruction, object orientation estimation, and marker tracking. The combination of the two technologies allow the Fin Ray finger to perform a greater variety of manipulation tasks, including the successful reorientation and placement of a transparent wine glass without any catastrophic failures. 

A few of the limitations are inherent to the twisting motion of the hollowed out finger, which is traded off with the rigidness of the acrylic at the back of the sensing region. This limitation caused the Fin Ray finger to require a nontrivial amount of force to comply to more curved objects. Additionally, if the object being grasped was too heavy or if the force was not enough to grab a smaller object to impart a large imprint on the silicone pad, the gripper would occasionally struggle with successful implementation of reconstruction and object orientation estimation. These issues can be fixed with future optimizations to the design of the hollowed out finger. 

Furthermore, although the tactile sensing software performed well most of the time, there were some issues which were caused from a lack of perfectly matched reference images. These issues could potentially be resolved with the addition of machine learning. Finally, the reconstruction images were uncalibrated and could not be used for a tactile sensing region height map. Although having a height map was not required for any of the tasks in this paper, future work could involve implementing a neural network to develop accurate height map reconstructions from the raw tactile images of the GelSight Fin Ray.  

Other future directions involve implementation of a GelSight Fin Ray into a 3-fingered gripper, which could leverage both the compliant grasping characteristics of the Fin Ray and its tactile sensing capabilities, to select and pick ripe fruit without bruising them, reducing food waste, or manipulating more dishes in a dishwasher placement task to work towards the development of a home-assistant robot that could allow the elderly to age with dignity.

\section{ACKNOWLEDGEMENTS}

Toyota Research Institute and the Office of Naval Research provided funds to support this work. The authors would also like to thank the members of the MIT International Design Center, Chris Haynes, Will Lutz, and Bill McKenna, for their help in manufacturing the finger; Alex Alspach for providing a fully 3D printed Fin Ray prototype; Branden Romero, Megha H. Tippur, Shaoxiong Wang, and Jerry Zhang for their helpful discussions that ranged from figures to coding to UR5 arms to design considerations; and finally, Alan Papalia for his assistance with ROS and machining.

\addtolength{\textheight}{-0cm}   




\bibliographystyle{IEEEtran}
\bibliography{Ref}
\end{document}